\def\year{2020}\relax
\documentclass[letterpaper]{article} 
\usepackage{aaai20}  
\usepackage{times}  
\usepackage{helvet} 
\usepackage{courier}  
\usepackage[hyphens]{url}  
\usepackage{graphicx} 
\usepackage{amsmath}
\usepackage{amssymb}
\usepackage{algorithm}
\usepackage{algorithmic}
\usepackage{graphicx}  
\title{Adversarial Transformations for Semi-Supervised Learning}
\author{Teppei Suzuki and Ikuro Sato\\
DENSO IT LABORATORY, INC.\\
 2-15-1 Shibuya, Shibuya-ku Tokyo, Japan\\
\{tsuzuki,isato\}@d-itlab.co.jp 
}

\begin{document}

\maketitle

\begin{abstract}
We propose a \textbf{R}egularization framework based on \textbf{A}dversarial \textbf{T}ransformations (RAT) for semi-supervised learning. RAT is designed to enhance robustness of the output distribution of class prediction for a given data against input perturbation. RAT is an extension of Virtual Adversarial Training (VAT) in such a way that RAT adversraialy transforms data along the underlying data distribution by a rich set of data transformation functions that leave class label invariant, whereas VAT simply produces adversarial additive noises. In addition, we verified that a technique of gradually increasing of perturbation region further improves the robustness. In experiments, we show that RAT significantly improves classification performance on CIFAR-10 and SVHN compared to existing regularization methods under standard semi-supervised image classification settings.
\end{abstract}

\section{Introduction}
Semi-supervised learning (SSL)~\cite{ssl} is an effective learning framework on datasets that have large amounts of the data and few labels. In a practical situations, obtained datasets are often partially labeled, because labeling is more costly than collecting data in many cases. Thus, a powerful SSL framework that enhances the model performance is needed.

\begin{figure}[t]
\centering
\begin{tabular}{cc}
\begin{minipage}{0.5\hsize}
\centering
\includegraphics[width=1\hsize]{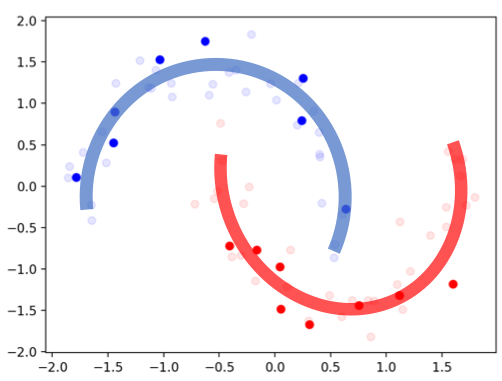}
\\ (a) input data
\end{minipage}
\begin{minipage}{0.5\hsize}
\centering
\includegraphics[width=1\hsize]{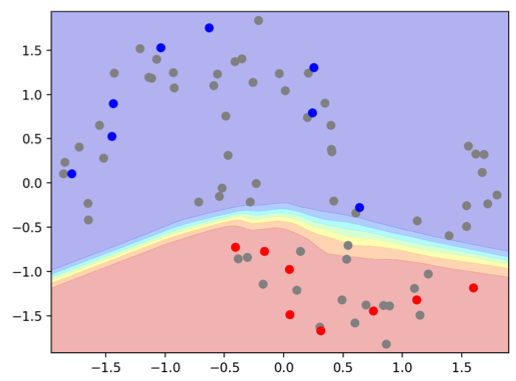}
\\ (b) VAT with $\epsilon=0.3$
\end{minipage}\\
\begin{minipage}{0.5\hsize}
\centering
\includegraphics[width=1\hsize]{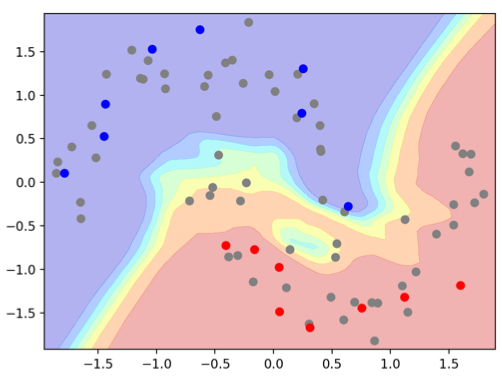}
\\ (c) VAT with $\epsilon=0.5$
\end{minipage}
\begin{minipage}{0.5\hsize}
\centering
\includegraphics[width=1\hsize]{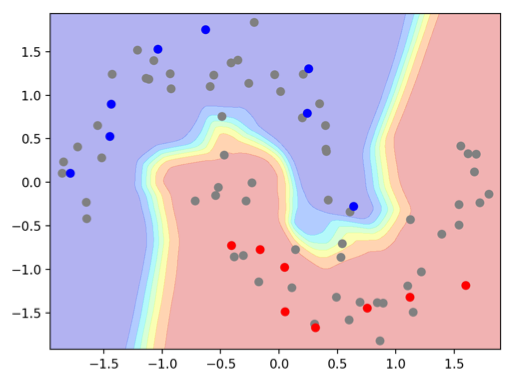}
\\ (d) RAT with rotation
\end{minipage}
\end{tabular}

\caption{Comparison between VAT and RAT with moons dataset. Colored regions of (b), (c), and (d) correspond to prediction confidence. Each moon has 10 labeled data (dark color points in (a)) and 30 unlabeled data (light color points in (a)), and each data is randomly generated from semicircles in (a) with Gaussian noise. RAT uses noise injection and rotation along the blue and red semicircles in (a) as transformation. The setting of (b) and (d) is the same except for using rotation in (d).}
\label{fig:concept image}
\end{figure}

Among the many SSL algorithms, virtual adversarial training (VAT)~\cite{vat2016,vat2018} is a successful one. It regularizes local distributional smoothness, by which we mean the robustness of the output distribution around each input datapoint against local perturbation, by using virtual adversarial perturbation. 
The perturbation is calculated as the noise, which adversarially changes the output of the model, and VAT imposes the consistency between the outputs of the model for the data and the perturbed data. It indicates that VAT enforces the output distribution to be robust with respect to perturbation within the $\epsilon$-ball centered on the data, where $\epsilon$ denotes the norm of the perturbation vector. The specific advantage of VAT is to leverage the adversarial perturbation.

Although VAT demonstrates remarkable accuracy in standard benchmarks, VAT allows the smoothing only within the $\epsilon$-ball. This means that there is no guarantee that two datapoints are classified into the same class even if they belong to the same class, if the distance between them is greater than $2\epsilon$ (see Figure 1 (b)). However, if $\epsilon$ is too large, the perturbed data may penetrate the true class boundary; inconsistency within $\epsilon$-ball may occur (see Figure 1 (c)). To summarize, the $\epsilon$-ball, that is an isotropic hypersphere in the input space, is too universal to take the underlying data distribution into account. Our basic idea is to use some adversarial transformations under which a datapoint transforms to another point within the underlying distribution of the same class.

We propose a regularization framework, RAT (\textbf{R}egularization based on \textbf{A}dversarial \textbf{T}ransformations), which regularizes the smoothness of the output distribution by utilizing adversarial transformations. The aim of RAT is to make the outputs with respect to the same class data close.

We justify RAT as the regularization for smoothing the distribution, and provide use of composite functions and a technique, $\epsilon$-rampup, that enlarge the area in which the smoothing is effective. To demonstrate the effectiveness of RAT, we compare it with baseline methods in the experiments following the valid setting proposed by Oliver et al. (2018), and RAT outperforms baseline methods.

We summarize our contributions as follows:
\begin{itemize}
    \item We propose the regularization framework based on adversarial transformations, which includes VAT as a special case. Unlike VAT, RAT can impose distributional smoothness along the underlying data distribution.
    \item We claim that use of composite transformations, each of which leaves class label invariant, can further improve the smoothing effect, because it enhances the degrees of freedom.
    \item Moreover, we provide a technique to enhance the smoothing effect by ramping up $\epsilon$. The technique is common to VAT and RAT, and enlarge the area in which the smoothing is effective.
    \item RAT outperforms baseline methods in semi-supervised image classification tasks. In particular, RAT is robust against reduction of labeled samples, compared to other methods.
\end{itemize}

\section{Virtual Adversarial Training: A Review}
\noindent In this section, we review virtual adversarial training (VAT)~\cite{vat2016,vat2018}. VAT is a similar method of the adversarial training~\cite{adv_exm_ian}, but the aim of VAT is to regularize distributional smoothness. Miyato et al. (2016; 2018) claimed importance of the adversarial noise and proved this by comparing it with random perturbation training. VAT has indeed shown state-of-the-art results in valid benchmarks~\cite{realisticeval}.

Let $x\in\mathbb{R}^I$ be a data sample where $I$ is the dimension of the input, and $p_\theta(y|x)$ be a model distribution parameterized by $\theta$. The objective function of VAT in SSL scenario is:
\begin{align}
    \mathcal{L}_\mathrm{VAT}(x_l,y,x_u,\theta)\equiv\mathcal{L}(x_l,y,\theta)-\lambda\mathrm{LDS}(x_u,\theta),
\end{align}
where $\mathcal{L}$ denotes a supervised loss term. $x_l,y$ and $x_u$ are labeled data, its labels, and unlabeled data. $\lambda$ is a scaling parameter for regularization. $\mathrm{LDS}$, local distributional smoothness, is defined as follows:
\begin{align}
    \mathrm{LDS}(x,\theta)\equiv-\Delta_\mathrm{KL}(r_\mathrm{v\mathchar`-adv},x,\theta),
\end{align}
where $\Delta_\mathrm{KL}$ and $r_\mathrm{v\mathchar`-adv}$ are KL divergence between output distributions with respect to an original data and a noise-added data, and virtual adversarial perturbation, respectively. They are defined as follows:
\begin{align}
    &\Delta_\mathrm{KL}(r,x,\theta)\equiv\mathrm{KL}\left[p_\theta(y|x)\| p_\theta(y|x+r)\right],\\ \label{eq:maximization_of_kl}
    &r_\mathrm{v\mathchar`-adv}\equiv\underset{r}{\mathrm{arg}\max}\{\Delta_\mathrm{KL}(r,x,\theta);\:\|r\|_2\leq\epsilon\},
\end{align}
where $\mathrm{KL}[\cdot\|\cdot]$ denotes KL divergence and $\|\cdot\|_2$ denotes the L2 norm. $\epsilon$ is a hyperparameter to control the range of the smoothing.

$r_\mathrm{v\mathchar`-adv}$ emerges as the eigenvector of Hessian matrix $H(x,\theta)\equiv\nabla\nabla_r\Delta_\mathrm{KL}(r,x,\theta)|_{r=0}$ from the following derivation. Since $\Delta_\mathrm{KL}(r,x,\theta)$ takes a minimum value $0$ at $r=0$, $\nabla_r\Delta_\mathrm{KL}(r,x,\theta)|_{r=0}$ is 0. Therefore, the second-order Taylor approximation of $\Delta_\mathrm{KL}(r,x,\theta)$ around $r=0$ is:
\begin{align}
    \label{eq:vat_hess}
    \Delta_\mathrm{KL}(r,x,\theta)\approx\frac{1}{2}r^\top H(x,\theta)r.
\end{align}
We can describe (\ref{eq:maximization_of_kl}) using (\ref{eq:vat_hess}) as follows:
\begin{align}
r_\mathrm{v\mathchar`-adv}\approx\underset{r}{\mathrm{arg}\max}\{r^\top H(x,\theta)r;\:\|r\|_2\leq\epsilon\}.
\end{align}
By this approximation, $r_\mathrm{v\mathchar`-adv}$ is parallel to the eigenvector corresponding to the largest eigenvalue of $H(x,\theta)$.
Note that VAT assumes that $p_\theta(y|x)$ is differentiable with respect to $\theta$ and $x$ almost everywhere, and we also assume it in this paper.

\section{Regularization Based on Adversarial Transformations}
We propose the regularization framework, RAT, which imposes the consistency between output distributions with respect to datapoints belonging to the same class. Leveraging the  power of adversariality, we introduce adversarial transformations that replace additive adversarial noises $r_\mathrm{v\mathchar`-adv}$ in VAT.

To consider imposing the consistency, we assume that each datapoint belonging to $k$-th class is in a class-specific subspace $\mathcal{S}_k$. We consider the generic transformation parameterized by $\phi$: for any $k$, $f_\phi:\mathcal{S}_k\rightarrow\mathcal{S}_k$ such that $\|\phi\|\leq\epsilon$, where $\|\cdot\|$ denotes valid norm with respect to $f_\phi$ such as L1, L2, or operator norm. Our strategy is to regularize the output distribution by utilizing $f_\phi$ instead of $r_\mathrm{v\mathchar`-adv}$. Note that we mainly consider the image classification tasks in this work. Therefore, we deal with the transformation such as the spatial transformation or the color distortion.

We define local distributional smoothness utilizing a transformation as follows:
\begin{align}
    \mathrm{LDS}_T(x,\theta)\equiv-\tilde{\Delta}_\mathrm{KL}(f_{\phi_\mathrm{T\mathchar`-adv}},x,\theta),
\end{align}
where $\tilde{\Delta}_\mathrm{KL}$ and $\phi_\mathrm{T\mathchar`-adv}$ are defined as follows:
\begin{align}
    \label{eq:tilde_KL}
    \tilde{\Delta}_\mathrm{KL}(f_\phi,x,\theta)\equiv\mathrm{KL}\left[p_\theta(y|x)\|p_\theta(y|f_\phi(x))\right],\\ \label{eq:argmax_tadv}
    \phi_\mathrm{T\mathchar`-adv}\equiv\underset{\phi}{\mathrm{arg}\max}\{\tilde{\Delta}_\mathrm{KL}(f_\phi,x,\theta);\:\|\phi\|\leq\epsilon\}.
\end{align}
We refer to $f_{\phi_\mathrm{T\mathchar`-adv}}$ as an adversarial transformation in this paper. There are some adversarial attacks utilizing functions instead of additive noises~\cite{adef,fourier_attack}, and the relation between adversarial transformation and these attacks is similar to the relation between virtual adversarial perturbation and adversarial perturbation~\cite{adv_exm_ian}.

We utilize $\mathrm{LDS}_T$ for imposing the consistency. Thus, the objective function of RAT for SSL scenario is represented as follows:
\begin{align}
    \mathcal{L}_\mathrm{RAT}(x_l,y,x_u,\theta)\equiv\mathcal{L}(x_l,y,\theta)-\lambda\mathrm{LDS}_T(x_u,\theta).
\end{align}
We can identify VAT as the special case of RAT; when $f_\phi(x)=x+\phi$ and $\|\cdot\|$ is the L2 norm, RAT is equal to VAT.

To compute $\mathrm{LDS}_T$, we have to solve the maximization problem (\ref{eq:argmax_tadv}). However, it is difficult to exactly solve it. Therefore, we consider approximating it by using Taylor approximation.

To efficiently approximate and solve (\ref{eq:argmax_tadv}), $f_\phi$ needs to satisfy the following two conditions:
\begin{itemize}
    \item[C1.] $f_\phi$ is differentiable with respect to $\phi$ almost everywhere.
    \item[C2.] There is a parameter $\phi_\mathrm{id}$ that makes $f_\phi$ identity transformation, $f_{\phi_\mathrm{id}}(x)=x$.
\end{itemize}
If $f_\phi$ satisfies these conditions, $\tilde{\Delta}_\mathrm{KL}(f_\phi,x,\theta)$ takes a minimum value at $\phi=\phi_\mathrm{id}$ and (\ref{eq:tilde_KL}) is written in the same form as (\ref{eq:vat_hess}) by the second-order Taylor approximation around $\phi_\mathrm{id}$ as  $\tilde{\Delta}_\mathrm{KL}(f_\phi,x,\theta)\approx\frac{1}{2}\phi^\top H(f_\phi, x,\theta)\phi$, where $H(f_\phi,x,\theta)\equiv\nabla\nabla_\phi\tilde{\Delta}_\mathrm{KL}(f_\phi,x,\theta)|_{\phi=\phi_\mathrm{id}}$. Thus, (\ref{eq:argmax_tadv}) is approximated as follows:
\begin{align}
    \label{eq:approx_tadv}
    \phi_\mathrm{T\mathchar`-adv}\approx\underset{\phi}{\mathrm{arg}\max}\{\phi^\top H(f_\phi,x,\theta)\phi;\:\|\phi\|\leq\epsilon\}.
\end{align}
$\phi_{T\mathrm{\mathchar`-adv}}$ is also parallel to the eigenvector corresponding to the largest eigenvalue of $H(f_\phi,x,\theta)$.

$f_\phi$ allows any transformations satisfying $f_\phi:\mathcal{S}_k\rightarrow\mathcal{S}_k$, and the conditions C1 and C2. In this paper, we use hand crafted transformations depending on input data domain such as image coordinate shift, image resizeing, or global color change, to name a few for the case of an image classification task. In a next section, we propose use of composite transformations as $f_\phi$ to further enhance the effect of the smoothing.

\subsection{Use of Composite Transformations}
Let $\mathcal{F}$ be a set of functions, $\{f_\phi:\mathcal{S}_k\rightarrow\mathcal{S}_k;\|\phi\|\leq\epsilon\}$, satisfying the conditions C1 and C2. $\mathcal{F}$ may contain composite functions of the form $\tilde{f}_{\tilde{\phi}}=f^{(1)}_{\phi^{(1)}} \circ f^{(2)}_{\phi^{(2)}} \cdots \circ f^{(n)}_{\phi^{(n)}}$, where $f^{(i)}_{\phi^{(i)}} \in \mathcal{F};\|\phi^{(i)}\|\leq\epsilon^{(i)}$ and $\tilde{\phi}\equiv\{\phi^{(i)}\}_{i=1}^n$. It is obvious that these composite functions $\tilde{f}_{\tilde{\phi}}: \mathcal{S}_k \rightarrow \mathcal{S}_k (\forall k)$ satisfy the conditions C1 and C2.

By having such composite functions, one can obtain a much richer set of transformations that still yield class-invariant transformations, as given by the relation, $\tilde{f}_{\tilde{\phi}}=f^{(j)}_{\phi^{(j)}}$ when $f^{(i)}_{\phi^{(i)}}=f^{(i)}_\mathrm{id}$ for all $i$ except for $j$. It is reasonable for RAT to utilize the composite functions, because the composite functions leads to the richer transformation and imposing the consistency over a wider range.

\subsection{Fast Approximation of $f_{\phi_\mathrm{T\mathchar`-adv}}$}
Although $\phi_{T\mathrm{\mathchar`-adv}}$ emerges as the eigenvector corresponding to the largest eigenvalue of Hessian matrix $\tilde{H}=\nabla\nabla_{\tilde{\phi}}\tilde{\Delta}_\mathrm{KL}(\tilde{f}_{\tilde{\phi}},x,\theta)|_{\tilde{\phi}=\tilde{\phi}_\mathrm{id}}$ as already described, the computational costs of the eigenvector are $O(n^3)$.
There is a way to approximate the eigenvector with small computational costs~\cite{power_iteration,vat2016,vat2018}.

We approximate the parameters $\tilde{\phi}$ with the power iteration and the finite difference method, just as VAT does. For each transformation $f^{(i)}\in\mathcal{F}$, we sample random unit vectors $\tilde{v}\equiv\{v^{(i)}\}_{i=1}^n$ as initial parameters and calculate $\tilde{v}\leftarrow\tilde{H}\tilde{v}$ iteratively. It makes $\tilde{v}$ converge to the eigenvector corresponding to the largest eigenvalue of $\tilde{H}$.

The Hessian-vector product is calculated with the finite difference method as follows:
\begin{align}
    \nonumber
    \tilde{H}\tilde{v}\approx&\frac{\nabla_{\tilde{\phi}}\tilde{\Delta}_\mathrm{KL}(\tilde{f}_{\tilde{\phi}},x,\theta)|_{\tilde{\phi}=\tilde{\phi}_\mathrm{id}+\xi\tilde{v}}-\nabla_{\tilde{\phi}}\tilde{\Delta}_\mathrm{KL}(\tilde{f}_{\tilde{\phi}},x,\theta)|_{\tilde{\phi}=\tilde{\phi}_\mathrm{id}}}{\xi},\\
    =&\frac{\nabla_{\tilde{\phi}}\tilde{\Delta}_\mathrm{KL}(\tilde{f}_{\tilde{\phi}},x,\theta)|_{\tilde{\phi}=\tilde{\phi}_\mathrm{id}+\xi\tilde{v}}}{\xi},
\end{align}
with $\xi\neq 0$. We used the fact that  $\nabla_{\tilde{\phi}}\tilde{\Delta}_\mathrm{KL}(\tilde{f}_{\tilde{\phi}},x,\theta)|_{\tilde{\phi}=\tilde{\phi}_\mathrm{id}}=0$. We approximate $\phi_{T\mathrm{\mathchar`-adv}}$ by normalizing the norm of the approximated eigenvector as described in the next section. Note that we calculate just one iteration for power iteration in this paper, because it is reported in \cite{vat2016,vat2018} that one iteration is sufficient for computing accurate $\tilde{H}\tilde{v}$ and increasing the iteration does not have an effects.

\subsection{$\epsilon$-Rampup}
Although $\phi_{T\mathrm{\mathchar`-adv}}$ should satisfy $\|\phi\|\leq\epsilon$, $\phi_{T\mathrm{\mathchar`-adv}}$ is actually given as the parameter satisfying $\|\phi\|=\epsilon$ when solving (\ref{eq:approx_tadv}), because the Hessian of KL divergence is semi-positive definite.

In the case of VAT, the smoothing with $\|r\|_2=\epsilon$ means that the model should satisfy consistency between the outputs with respect to the original data and the data on the surface of the $\epsilon$-ball centered on the original data.

We propose a technique, $\epsilon$-rampup, that enhances the effect of the smoothing not only on the boundary but also inside the boundary with small computational costs. The technique is to ramp up $\epsilon$ from 0 to a predefined value during training, and the parameters of the adversarial transformation are determined by solving $\mathrm{arg}\max_\phi\{\tilde{\Delta}_\mathrm{KL}(f_\phi,x,\theta);\|\phi\|=\epsilon\}$. One can approximately solve this by normalizing the approximated eigenvector to satisfy $\|\phi\|=\epsilon$ after the power iteration.

Although there are many techniques to ramp up or anneal parameters~\cite{meanteacher,cyclr}, we adopt the procedure used in Mean Teacher~\cite{meanteacher}. Mean Teacher utilizes a sigmoid shape function, $\exp(-5(1-x)^2);\:x=[0,1]$, for ramping up the regularization coefficient, and we adopt it for $\epsilon$.

We show the pseudocode of the generation process of $\phi_{T\mathrm{\mathchar`-adv}}$ in Algorithm \ref{alg:avat}.

\begin{algorithm}[t]
\caption{Generation of $\phi_{T\mathrm{\mathchar`-adv}}$}
\label{alg:avat}
\begin{algorithmic}[1]
\STATE \textbf{Input:} Data $x\in\mathcal{X}$; transformation functions $\{f^{(i)}\}_{i=1}^n\in\mathcal{F}$; scalar parameters $\{\epsilon^{(i)}\}_{i=1}^n$ and $\xi$
\STATE \textbf{Output:} function parameters $\{\phi^{(i)}_{T\mathrm{\mathchar`-adv}}\}_{i=1}^n$
\STATE Make copy of data $\hat{x}\leftarrow x$
\FOR{$i=1,\dots, n$}
\STATE initialize $v^{(i)}$ as a random unit vector
\STATE $\phi^{(i)}=\phi^{(i)}_\mathrm{id}+\xi v^{(i)}$
\STATE $\hat{x}\leftarrow f^{(i)}_{\phi^{(i)}}(\hat{x})$
\ENDFOR
\FOR{$i=1,\dots,n$}
\STATE $v^{(i)}\leftarrow\nabla_{v^{(i)}}\mathrm{KL}[p_\theta(y|x)\|p_\theta(y|\hat{x})]$
\STATE $\epsilon^{(i)}\leftarrow\mathrm{Rampup}(\epsilon^{(i)})$
\STATE Normalize $v^{(i)}$ to satisfy $\|\phi^{(i)}\|=\epsilon^{(i)}$
\STATE $\phi_{T\mathrm{\mathchar`-adv}}^{(i)}=\phi^{(i)}_\mathrm{id}+v^{(i)}$
\ENDFOR
\end{algorithmic}
\end{algorithm}
\subsection{Evaluation on Synthetic Dataset}
We show the smoothing effect of RAT with a toy problem, a moons dataset, in Figure \ref{fig:concept image}. We make 10 labeled samples and 30 labeled samples for each moon. $p_\theta(y|x)$ consists of a three-layer neural network with ReLU non-linearity~\cite{relu}. All hiden layers have 128 units. $\theta$ is optimized with Adam optimizer for 500 iterations with default parameters suggested in \cite{adam}. In each iteration, we use all samples for updating $\theta$. We treat this toy problem as if we already know appropriate class-invariant transformations; we adopt class-wise rotation along each moon as $\tilde{f}_{\tilde{\phi}}(x)=R(x,\phi^{(1)})+\phi^{(2)};\|\phi^{(2)}\|_2\leq0.3,|\phi^{(1)}|\leq10^\circ$ for illustraintion purpose. Note that we do not ramp up $\epsilon$ in this experiment.

VAT with a small $\epsilon$ draws the decision boundary crossing $\mathcal{S}_k$ as shown in Figure \ref{fig:concept image} (b). When we adopt a lager $\epsilon$, VAT cannot smooth the output distribution within the $\epsilon$-ball as shown in Figure \ref{fig:concept image} (c), because the larger $\epsilon$ allows the unexpected transformation $f:\mathcal{S}_k\rightarrow\mathcal{S}_{k'}$, and causes inconsistency. On the other hand, RAT draws the decision boundary along $\mathcal{S}_k$.

In this toy problem, we utilize $\mathcal{S}_k$ and it is equal to using the label information implicitly. Therefore, in a next section, we evaluate RAT using realistic situations, where we do not know $\mathcal{S}_k$.

\section{Experiments}
We evaluate the effectiveness of RAT and $\epsilon$-rampup through three experiments on a semi-supervised image classification task: (i) evaluation of composite transformations, (ii) evaluation of $\epsilon$-rampup for VAT and RAT, and (iii) comparison of RAT to baseline methods. As the baseline methods, we use $\Pi$-Model~\cite{pimodel1,pimodel2}, Pseudo-label~\cite{pseudo_label}, Mean Teacher~\cite{meanteacher}, and VAT~\cite{vat2018}. Note that VAT utilizes entropy regularization in all experiments. To evaluate on realistic situations, we follow the setting proposed by Oliver et al. (2018). We use PyTorch~\cite{pytorch} to implement and evaluate SSL algorithms, and we carefully reproduced the results of Oliver et al. (2018).
All hyperparameters for SSL algorithms are adopted the same as in Oliver et al. (2018) except that we do not use L1 and L2 regularization. 

For all experiments, we used the same Wide ResNet architecture, depth 28 and width 2~\cite{wrn}, and we use the CIFAR-10~\cite{cifar} and SVHN~\cite{svhn} datasets for evaluation. CIFAR-10 has 50,000 training data and 10,000 test data, and we split training data into a train/validation set, 45,000 data for training and 5,000 data for validation. SVHN has 73,257 data for training and 26,032 data for testing. We also split training data into 65,931 data for training and 7,326 data for validation. For the semi-supervised setting, we further split training data into labeled data and unlabeled data.

We utilize standard preprocessing and data augmentations for training, following Oliver et al. (2018). For SVHN, we normalize the pixel value into the range $[-1,1]$, and use random translation by up to 2 pixels as data augmentation. For CIFAR-10, we apply ZCA normalization~\cite{cifar} and global contrast normalization as normalization, and random horizontal flipping, random translation by up to 2 pixels, and Gaussian noise injection with standard deviation 0.15 as data augmentation.

We report the mean and standard deviation of error rates over five trials with test sets. The test error rates are evaluated with the model that has the minimum error rate on validation sets. The evaluation on validation set is executed every 25,000 training iterations.

\subsection{Implementation Details of RAT}
We tested three types of data transformations, all of which are commonly used in data augmentation in image classification tasks. All transformations $f_\phi$ discussed below satisfy the conditions C1 and C2. We evaluated different types of composite transformations as is discussed below.

\subsubsection{Noise Injection}
The noise injection is represented as $f_\phi(x)=x+\phi$. We define the norm for the parameters of the noise injection as the L2 norm, $\|\phi\|_2$. This is equal to the formulation of VAT.

\subsubsection{Spatial Transformation}
We consider three spatial transformations with different degrees of freedom: affine transformation, thin plate spline~\cite{tps}, and flow field. All these transformations shift the pixel position $(u,v)$ by offset vector $(\delta u, \delta v)$ to give $(u+\delta u,v+\delta v)$. The pixel values of the transformed image are calculated by bilinear interpolation. The details of these transformations are provided in \cite{stn}.

The difference between transformations is the degrees of freedom to calculate the offset vectors as follows: affine transformation has six parameters, thin plate spline has parameters proportional to the number of control points $m$, and the flow field directly has the local offset vectors as parameters, meaning that the number of parameters of the flow field is proportional to the spatial resolution of the image. We set $m$ to $16$ in all experiments, which means we employ a $4\times 4$ grid.

We define the norm for the parameters of the flow field transformations as the L2 norm, $\sqrt{\sum_i^N(\delta u^2_i+\delta v^2_i)}$, where $N$ is the number of pixels. The norm for thin plate spline is also defined as the L2 norm of offset vectors for the control points.

Affine transformation is a linear operator in homogeneous coordinates, and the norm is given as the operator norm. Thus, $\|\phi\|$ is calculated as the maximum singular value of an affine transformation matrix.

\subsubsection{Color Distortion}
Color distortion is an effective augmentation method for image classification tasks. Among many methods for color distortion, we use a simple way, channel-wise weighting, $f_\phi(x_c)=\phi_c x_{c,i}$, where $x_{c,i}$ is the pixel value of the $c$-th channel of the $i$-th pixel, and $\phi_c$ is the scalar value for each channel. This transformation is described as the linear operator, and we define the norm as the operator norm. Note that channel-wise weighting is represented as the multiplication of a diagonal matrix and a pixel value, and the operator norm is calculated as $\max_c|\phi_c|$.

\subsection{Evaluation of Composite Transformations}
Since the performance of RAT depends on the combination of the transformations, we report the effect of a combination of functions by adding transformations to VAT.

We first seek good $\epsilon$ for each transformation with a grid search on CIFAR-10 with 4,000 labeled data, from 0.001 to 0.01 with 0.001 step size for channel-wise weighting, from 0.1 to 1 with a 0.1 step size for affine transformation and thin plate spline, and from 0.01 to 0.1 with a 0.01 step size for flow field. We show the grid search results in Table \ref{tab:hyper_param}. Other parameters such as $\xi$, $\lambda$, and parameters for optimization are the same as VAT suggested in \cite{realisticeval}. We summarize the parameters in Table \ref{tab:vat_rat_param}. Note that $\epsilon$-rampup is not utilized in this experiment, because this experiment explores the effect of composite transformations.

\begin{table}[t]
\centering
\caption{The parameters $\epsilon$ for transformations.}
\label{tab:hyper_param}
\begin{tabular}{cc}
\hline
Transformations        & $\epsilon$ \\ \hline
Channel-wise weighting & 0.001    \\
Affine transformation  & 0.6     \\
Thin plate spline      & 1       \\
Flow field             & 0.01   \\ \hline
\end{tabular}
\end{table}
\begin{table}[t]
\centering
\caption{VAT and RAT shared parameters. $\epsilon$ is required 6.0 for CIFAR-10 and 1.0 for SVHN.}
\label{tab:vat_rat_param}
\begin{tabular}{cc}
\hline
Parameters & values \\ \hline
Initial learning rate       & 0.003                    \\
Max consistency coefficient $\lambda$ & 0.3                      \\
noise injection $\epsilon$ & 6.0 or 1.0               \\
$\xi$          & $10^{-6}$                     \\
Entropy penalty multiplier  & 0.06 \\ \hline
\end{tabular}
\end{table}

The results of adopting various transformations with CIFAR-10 with 4,000 labeled data and SVHN with 1,000 labeled data are shown in Table \ref{tab:ablation_study}. 
\begin{table*}[t]
\centering
\caption{Test error rates of RAT with various transformations on CIFAR-10 with 4,000 labeled data and SVHN with 1,000 labeled data. All settings of RAT include noise injection as transformation, and all of the hyperparameters and experiment settings of VAT and RAT are the same, except for the inherent parameters of RAT.}
\label{tab:ablation_study}
\begin{tabular}{l|cccc|cc}
\hline
Methods & Channel-wise & Affine & Thin Plate Spline & Flow Field & CIFAR-10 & SVHN \\ \hline
Supervised & & & & & 20.35$\pm$0.14\% & 12.33$\pm$0.25\% \\ \hline
VAT  &        &     &          &         & 13.68$\pm$0.25\% & 5.32$\pm$0.25\% \\ \hline
RAT & \checkmark  &       &        &           & 14.33$\pm$0.44\% &5.19$\pm$0.29\% \\ 
     & &  \checkmark  &        &           &11.70$\pm$0.32\%  &\textbf{3.10$\pm$0.12\%} \\ 
     & & &  \checkmark  &        &  13.06$\pm$0.44\% & 4.12$\pm$0.11\% \\ 
     & & & &  \checkmark  & 14.27$\pm$0.41\% & 4.93$\pm$0.35\%  \\ 
     & \checkmark & \checkmark & &      & \textbf{11.32$\pm$0.44}\% & 3.14$\pm$0.12\%\\ 
     & \checkmark & & \checkmark  &  & 13.46$\pm$0.37\% & 4.06$\pm$0.16\% \\ 
     & \checkmark & & & \checkmark & 14.17$\pm$0.27\% & 4.85$\pm$0.12\% \\ \hline 
\end{tabular}
\end{table*}
All transformations and combinations improve the performance from supervised learning. However, channel-wise weighting and flow field increase the test error rates from VAT in CIFAR-10. For CIFAR-10, since we apply the ZCA normalization and the global contrast normalization, channel-wise weighting for the space of normalized data is an unnatural transformation for natural images. On the other hand, flow field is the transformation that can break object structure. Unlike a simple structure like the data in SVHN, the detail structure is the important feature for general objects. Thus, flow field induces the unfavorable effect.

Affine transformation achieves the best performance of all spatial transformations, and the results are induced by the low degree of freedom of the affine transformation. The affine transformation has the lowest degree of freedom among the three. In particular, the affine transformation preserves points, straight lines and planes. In other words, the affine transformation preserves the basic structure of the objects. Therefore, except for extreme cases, the affine transformation ensures that the class of transformed data is the same as the class of original data. This fact matches the strategy of RAT, which is that the output distribution of the data belonging to the same class should be close.
\begin{figure*}[t]
    \centering
    \begin{tabular}{cc}
    \begin{minipage}{0.5\hsize}
    \centering
    \includegraphics[width=1.0\hsize]{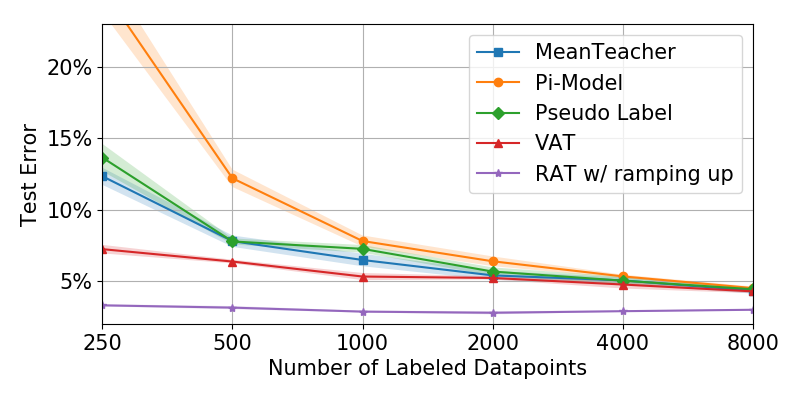}
    (a) SVHN
    \end{minipage}
    \begin{minipage}{0.5\hsize}
    \centering
    \includegraphics[width=1.0\hsize]{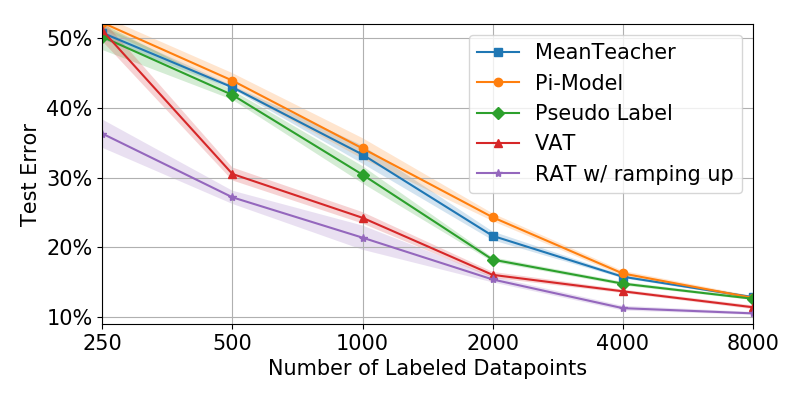}
    (b) CIFAR-10
    \end{minipage}
    \end{tabular}
    \caption{Test error rates obtained by varying the number of labeled data. Shaded regions indicate standard deviation over five trials.}
    \label{fig:varying_number}
\end{figure*}

The effect of combining channel-wise weighting and each spatial transformation is less effective. This fact means that combining channel-wise weighting does not expand the smoothing effect to a meaningful range. Indeed, the difference of combining channel-wise weighting is within the standard deviation. 

\subsection{Evaluation of $\epsilon$-Rampup}
We evaluate the effectiveness of $\epsilon$-rampup with CIFAR-10 with 4,000 labeled data. Since $\epsilon$-rampup is the technique for VAT and RAT, we compare the results with and without ramping up for VAT and RAT. We ramp up $\epsilon$ for 400,000 iterations. We utilize the composite transformations consisting of affine transformation and noise injection for RAT, and the hyperparameters are as in Table \ref{tab:vat_rat_param}.

The results are shown in Table \ref{tab:rampingup}. For both VAT and RAT, $\epsilon$-rampup brings a positive effect. As interesting effects, $\epsilon$-rampup allows for a large $\epsilon$. Since the smoothing effect reaches within the range of $\|\phi\|\leq\epsilon$ by ramping up, VAT and RAT with ramping up work well with a relatively large $\epsilon$.
\begin{table}[t]
\centering
\caption{Comparison between with and without $\epsilon$-rampup on CIFAR-10 with 4,000 labeled data. RAT has two $\epsilon$, one for the noise injection and one for the affine transformation.}
\label{tab:rampingup}
\begin{tabular}{l|c|c}
\hline
Methods & maximum $\epsilon$ & CIFAR-10          \\ \hline
VAT & 6 & 13.68$\pm$0.25\% \\
VAT & 10 & 14.31$\pm$0.33\% \\
VAT w/ $\epsilon$-rampup & 10 & \textbf{13.26$\pm$0.20\%} \\ \hline
RAT & (6, 0.6) & 11.70$\pm$0.32\% \\
RAT & (6, 1) & 12.68$\pm$0.29\% \\
RAT w/ $\epsilon$-rampup & (6, 1) & \textbf{11.26$\pm$0.34\%}  \\ \hline
\end{tabular}
\end{table}

\subsection{Comparison of RAT to Baselines}
We show the effectiveness of RAT by comparing it with baseline methods and non-adversarial version of RAT (Random Transformation). We use the composite transformations consisting of affine transformation and noise injection for RAT. The hyperparameters of RAT and $\epsilon$ for transformations are the same as shown in Tables \ref{tab:hyper_param} and \ref{tab:vat_rat_param}, respectively, except that we set $\epsilon$ for affine transformation to 1 for RAT with ramping up with CIFAR-10, and set $\epsilon$ for noise injection to 5 for RAT and VAT with $\epsilon$-rampup with SVHN. Note that all the parameters of the random transformation are the same as RAT.

In Table \ref{tab:standard_comparison}, we show the comparison results with standard SSL settings, CIFAR-10 with 4,000 labeled data and SVHN with 1,000 labeled data.
\begin{table}[t]
\centering
\caption{Test error rates of RAT and baseline methods on CIFAR-10 with 4,000 labeled data and SVHN with 1,000 labeled data.}
\label{tab:standard_comparison}
\begin{tabular}{l|cc}
\hline
             & CIFAR-10      & SVHN          \\
Methods      & 4,000 Labels   & 1,000 Labels    \\ \hline
Supervised   & 20.35$\pm$0.14\% & 12.33$\pm$0.25\% \\
$\Pi$-Model  & 16.24$\pm$0.38\% & 7.81$\pm$0.39\%  \\
Pseudo-Label & 14.78$\pm$0.26\% & 7.26$\pm$0.27\%  \\
Mean Teacher & 15.77$\pm$0.22\% & 6.48$\pm$0.44\%  \\
VAT          & 13.68$\pm$0.25\% & 5.32$\pm$0.25\%  \\
VAT w/ $\epsilon$-rampup & \textbf{13.26$\pm$0.20\%} & \textbf{5.17$\pm$0.26\%}  \\ \hline
Random Transformation & 12.71$\pm$0.82\% & 6.06$\pm$0.50\% \\
RAT          & 11.70$\pm$0.32\% & 3.10$\pm$0.12\%   \\
RAT w/ $\epsilon$-rampup & \textbf{11.26$\pm$0.34\%} & \textbf{2.86$\pm$0.07\%} \\ \hline
\end{tabular}
\end{table}
On both datasets, RAT improves test error rates more than 2\% from the best baseline method, VAT. Futhermore, RAT also improves the error rates from the random transformation. The results prove the importance of the adversariality.

We evaluate the test error rates of each method with a varying number of labeled data from 250 to 8,000. The results are shown in Figure \ref{fig:varying_number}. RAT consistently outperforms other methods for all the range on both datasets. Remarkably, RAT significantly improves the test error rate in CIFAR-10 with 250 labeled data compared with the best result of baseline methods, from 50.20$\pm$1.88\% to 36.31$\pm$2.03\%. We believe that the improvement results from appropriately smoothed model prediction along the underlying data distribution. The experimental fact of VAT's underperformance, namely 15\% degraded CIFAR-10 test error rate compared to RAT, is a clear indication that adversarial, class-invariant transformation provides far better consistency regularization than isotropic, adversarial noise superposition.

Lastly, we compare our results with very recently reported results of Mixup-based SSL method, MixMatch~\cite{mixmatch}. Interestingly,  RAT is comparative or superior to MixMatch on SVHN, while MixMatch is superior to RAT on CIFAR-10. But, we should point out that this comparison does not seem fair because the experimental settings of MixMatch are different in several respects from ours and others shown in Figure \ref{fig:varying_number}. In MixMatch paper, the authors took exponential moving average of models, and the final results were given by the median of the last 20 test error rates measured every $2^{16}$ training iterations. These settings, that are missing in our experiments, seem to partly boost their performance. Readers are referred to Appendix for the detailed comparison with MixMatch.

\section{Related Work}
There are many SSL algorithms such as graph-based methods~\cite{gcn,label_prop}, generative model-based methods~\cite{semi_vae,semi_gan,mgan}, and regularization-based methods~\cite{vat2016,pimodel1,pimodel2,meanteacher,mixmatch}.

Label propagation~\cite{label_prop} is a representative graph-based SSL method. Modern graph-based approaches utilize neural networks for graphs~\cite{gcn}. Graph-based methods demonstrate the effectiveness, when elaborated graph structure is given.

Generative models such as VAE and GAN are now popular frameworks for the SSL setting~\cite{semi_vae,semi_gan}. Although generative model-based SSL methods typically have to train additional models, they tend to show remarkable gain in test performance.

Regularization-based methods are comparatively much more tractable, and can be utilized in arbitrary models. Next, we review three regularization-based methods closely related to RAT.

\subsection{Consistency Regularization}
Consistency regularization is a method imposing the consistency between the outputs of one model with respect to a typically unlabeled data and its perturbed counterpart, or the outputs of two models with respect to the same input. One of the simplest ways of constructing consistency regularization is to add stochastic perturbation to the data, $x\rightarrow\hat{x}$, as follows:
\begin{align}
    \min_\theta d(p_\theta(y|x),p_\theta(y|\hat{x}))
\end{align}
where $d$ is some distance functions; e.g., Euclidean distance or KL divergence. VAT~\cite{vat2016,vat2018} and RAT are classified in this category.

The random transformation-based consistency regularization techniques, the work~\cite{pimodel1} and ``$\Pi$-Model''~\cite{pimodel2} are vary similar and famous ones. We refer to these models as $\Pi$-Model in this paper. One can view $\Pi$-Model as non-adversarial version of RAT.

The other way of constructing consistency regularization is to utilize dropout~\cite{dropout}. Let $\theta_1$ and $\theta_2$ be the randomly selected parameters through dropout. Dropout as consistency regularization is represented as follows:
\begin{align}
    \min_{\theta_1,\theta_2\sim\theta}d(p_{\theta_1}(y|x),p_{\theta_2}(y|x)).
\end{align}

Mean Teacher~\cite{meanteacher} is a successful method that employs consistency regularization between two models. It makes a teacher model by exponential moving average of the parameters of a student model, and imposes the consistency between the teacher and the student. Although Mean Teacher can be combined with other SSL methods, the combination sometimes impairs the model performance as reported in \cite{mixmatch}.

VAT~\cite{vat2016,vat2018} is a very effective consistency regularization method. The advantage of VAT lies in the generation of adversarial noises, and the adversariality leads to isotropic smoothness around sampled datapoints. VAT also show the effectiveness in natural language processing tasks~\cite{nvat}.

\subsection{Entropy Regularization}
Entropy regularization is a way to bring low entropy on $p_\theta(y|x)$ to make model prediction more discriminative, and is known to give low-density separation~\cite{entmin,ssl}. The entropy regularization term is:
\begin{align}
    \min_\theta-p_\theta(y|x)\log p_\theta(y|x).
\end{align}
This regularization is often combined with other SSL algorithms~\cite{pimodel1,vat2018} and a combined method, VAT+entropy regularization, shows the state-of-the-art results in \cite{realisticeval}.

\subsection{Mixup}
Mixup~\cite{mixup} is a powerful regularization method that is very recently used for SSL~\cite{manifold_mixup,mixmatch,ict}. Mixup blends two different data, $x_1$ and $x_2$, and their labels, $y_1$ and $y_2$, as follows:
\begin{align}
    \hat{x}=\beta x_1+(1-\beta)x_2,\:\hat{y}=\beta y_1+(1-\beta)y_2,
\end{align}
where $\beta$ is the scalar value sampled from Beta distribution. In a semi-supervised setting, $\hat{y}$ is calculated as a blend between a label and a prediction or predictions, $\beta y_1 + (1-\beta)p_\theta(y|x_2)$ or $\beta p_\theta(y|x_1)+(1-\beta)p_\theta(y|x_2)$, and a regularization term is described as consistency regularization, $\min_\theta d(\hat{y},p_\theta(y|\hat{x}))$.

\section{Conclusion}
We proposed an SSL framework, called RAT, Regularization framework based on Adversarial Transformation. RAT aims to smooth model output along the underlying data distribution within a given class based on recent advancement of generation of adversarial inputs that stem from unlabeled data. Instead of just superposing adversarial noise, RAT uses a wider range of data transformations, each of which leaves class label invariant. We further propose use of composite transformations and a technique, called $\epsilon$-rampup, to enlarge the area in which the smoothing is effective without sacrificing computational cost. We experimentally show that RAT significantly outperform the baseline methods including VAT in semi-supervised image classification tasks. RAT is especially robust against reduction of labeled samples, compared to other methods. As a future work, we would like to replace the designing of composite functions by black box function optimization.

\bibliographystyle{aaai}
\bibliography{references.bib}

\newpage
\onecolumn
\section{Appendix: Adversarial Transformations for Semi-Supervised Learning\\ ID: 3743}

\subsection{A. Hyperparemters}
Hyperparameters for all baseline methods except for VAT are shown in Table \ref{tab:sup_hyp}. VAT and RAT hyperparameters are given in the main text.
\begin{table}[h]
\centering
\caption{Hyperparemter settings. All hyperparameters were determined following Oliver et al. (2018) except for L1 and L2 regularization. In our experiments, we do not use L1 and L2 regularization, because the original implementation of Oliver et al. (2018) do not use them.}
\label{tab:sup_hyp}
\begin{tabular}{lc}
\hline
\multicolumn{2}{c}{Shared}                 \\ \hline
Training iteration               & 500,000 \\
Learning decayed by a factor of  & 0.2     \\
at training iteration            & 400,000 \\
Consistency coefficient rampup   & 200,000 \\ \hline
\multicolumn{2}{c}{Supervised}             \\ \hline
Initial learning rate            & 0.003   \\ \hline
\multicolumn{2}{c}{$\Pi$-Model}            \\ \hline
Initial learning rate            & 0.0003  \\
Max consistency coefficient      & 20      \\ \hline
\multicolumn{2}{c}{Mean Teacher}           \\ \hline
Initial learning rate            & 0.0004  \\
Max consistency coefficient      & 8       \\
Exponential moving average decay & 0.95    \\ \hline
\multicolumn{2}{c}{Pseudo-Label}           \\ \hline
Initial learning rate            & 0.003   \\
Max consistency coefficient      & 1.0     \\
Pseudo-label threshold           & 0.95   \\ \hline
\end{tabular}
\end{table}

\subsection{B. Comparison of RAT to MixMatch}
We compare a Mixup-based method, MixMatch~\cite{mixmatch} in Tables \ref{tab:comp_mixmatch_svhn} and \ref{tab:comp_mixmatch_cifar}. MixMatch is an unrefereed state-of-the-art method at the time of submission. MixMatch combines various semi-supervised learning (SSL) mechanisms, such as consistency regularization, pseudo-label, and mixup. As is mentioned in the main text, the comparisons in Tables \ref{tab:comp_mixmatch_svhn} and \ref{tab:comp_mixmatch_cifar} are not exactly fair. In MixMatch paper, the authors took exponential moving average of models, and the final results were given by the median of the last 20 test error rates measured every $2^{16}$ training iterations. These settings are missing in our experiments.

We show the results on SVHN, varying the number of labeled data in Table \ref{tab:comp_mixmatch_svhn}. MixMatch results are taken from \cite{mixmatch}.
Although the experiment setting is slightly different, RAT outperforms MixMatch except for the 4000 labeled setting. The test error rates of RAT increase from 2,000 labeled data to 8,000 labeled data settings. The hyperparameters of RAT were determined on CIFAR-10 with 4,000 labeled data and did not consider SVHN. Nevertheless, the test error rates of RAT on SVHN 4000 labeled are the same as MixMatch.

We also show the results on CIFAR-10 in Table \ref{tab:comp_mixmatch_cifar}. MixMatch demonstrated remarkable results.

Since RAT and Mixup are compatible, RAT has the potential to improve test error rates by Mixup. However, exploring the combination of SSL methods is outside the scope of this paper. We will evaluate RAT+Mixup in future work.
\begin{table}[h]
\centering
\caption{Test error rates (\%) for SVHN}
\label{tab:comp_mixmatch_svhn}
\begin{tabular}{r|cc}
\hline
labels & MixMatch      & RAT w/ $\epsilon$-rampup\\ \hline
250    & 3.78$\pm$0.26 & 3.30$\pm$0.08           \\
500    & 3.64$\pm$0.46 & 3.14$\pm$0.11           \\
1000   & 3.27$\pm$0.31 & 2.86$\pm$0.07           \\
2000   & 3.04$\pm$0.13 & 2.78$\pm$0.11           \\
4000   & 2.89$\pm$0.06 & 2.89$\pm$0.05           \\
8000   & N/A           & 2.99$\pm$0.05           \\ \hline
\end{tabular}
\end{table}
\begin{table}[h]
\centering
\caption{Test error rates (\%) for CIFAR-10}
\label{tab:comp_mixmatch_cifar}
\begin{tabular}{l|rr}
\hline
labels             & 250           & 4000         \\ \hline
MixMatch w/o Mixup & 39.11         & 10.97 \\
MixMatch           & 11.80         & 6.00 \\
RAT w/ $\epsilon$-rampup  & 36.31$\pm$2.03& 11.26$\pm$0.34 \\ \hline
\end{tabular}
\end{table}

\subsection{C. Test error rates of Figure \ref{fig:varying_number}}
We show the test error rates of Figure \ref{fig:varying_number} in Tables \ref{tab:varying_res_cifar} and \ref{tab:varying_res_svhn}.
\begin{table*}[t]
\centering
\caption{Test error rates (\%) for CIFAR-10}
\label{tab:varying_res_cifar}
\begin{tabular}{l|rrrrrr}
\hline
labels            & 250           & 500           & 1000          & 2000          & 4000          & 8000          \\ \hline
$\Pi$-Model       & 52.24$\pm$0.94 & 43.92$\pm$1.08 & 34.19$\pm$1.48 & 24.31$\pm$0.55 & 16.24$\pm$0.38 & 12.74$\pm$0.25 \\
Pseudo-label      & 50.20$\pm$1.88 & 41.86$\pm$0.68 & 30.42$\pm$1.24 & 18.23$\pm$0.33 & 14.78$\pm$0.26 & 12.61$\pm$0.19 \\
Mean Teacher      & 50.78$\pm$0.93 & 42.96$\pm$0.25 & 33.31$\pm$1.30 & 21.64$\pm$0.63 & 15.77$\pm$0.22 & 12.84$\pm$0.18 \\
VAT               & 51.12$\pm$1.54 & 30.52$\pm$0.93 & 24.23$\pm$0.82 & 16.04$\pm$0.43 & 13.68$\pm$0.25 & 11.39$\pm$0.27 \\
RAT w/ $\epsilon$-rampup & 36.31$\pm$2.03 & 27.17$\pm$0.95 & 21.38$\pm$1.74 & 15.38$\pm$0.42 & 11.26$\pm$0.34 & 10.51$\pm$0.21 \\ \hline
\end{tabular}
\end{table*}
\begin{table*}[t]
\centering
\caption{Test error rates (\%) for SVHN}
\label{tab:varying_res_svhn}
\begin{tabular}{l|rrrrrr}
\hline
labels            & 250           & 500           & 1000         & 2000         & 4000         & 8000         \\ \hline
$\Pi$-Model       & 25.51$\pm$1.67 & 12.21$\pm$0.61 & 7.81$\pm$0.39 & 6.39$\pm$0.34 & 5.33$\pm$0.12 & 4.51$\pm$0.11 \\
Pseudo-label      & 13.66$\pm$0.96 & 7.78$\pm$0.21  & 7.26$\pm$0.27 & 5.67$\pm$0.29 & 5.03$\pm$0.28 & 4.43$\pm$0.12 \\
Mean Teacher      & 12.35$\pm$0.60 & 7.80$\pm$0.40  & 6.48$\pm$0.44 & 5.40$\pm$0.32 & 5.05$\pm$0.21 & 4.33$\pm$0.21 \\
VAT               & 7.24$\pm$0.30  & 6.37$\pm$0.13  & 5.32$\pm$0.25 & 5.22$\pm$0.11 & 4.76$\pm$0.28 & 4.28$\pm$0.13 \\
RAT w/ $\epsilon$-rampup & 3.30$\pm$0.08  & 3.14$\pm$0.11  & 2.86$\pm$0.07 & 2.78$\pm$0.11 & 2.89$\pm$0.05 & 2.99$\pm$0.05 \\ \hline
\end{tabular}
\end{table*}

\end{document}